\RequirePackage{amsmath}
\documentclass[runningheads,a4paper]{llncs}\usepackage[]{graphicx}\usepackage[]{color}
\makeatletter
\def\maxwidth{ %
  \ifdim\Gin@nat@width>\linewidth
    \linewidth
  \else
    \Gin@nat@width
  \fi
}
\makeatother

\definecolor{fgcolor}{rgb}{0.345, 0.345, 0.345}
\newcommand{\hlnum}[1]{\textcolor[rgb]{0.686,0.059,0.569}{#1}}%
\newcommand{\hlstr}[1]{\textcolor[rgb]{0.192,0.494,0.8}{#1}}%
\newcommand{\hlcom}[1]{\textcolor[rgb]{0.678,0.584,0.686}{\textit{#1}}}%
\newcommand{\hlopt}[1]{\textcolor[rgb]{0,0,0}{#1}}%
\newcommand{\hlstd}[1]{\textcolor[rgb]{0.345,0.345,0.345}{#1}}%
\newcommand{\hlkwa}[1]{\textcolor[rgb]{0.161,0.373,0.58}{\textbf{#1}}}%
\newcommand{\hlkwb}[1]{\textcolor[rgb]{0.69,0.353,0.396}{#1}}%
\newcommand{\hlkwc}[1]{\textcolor[rgb]{0.333,0.667,0.333}{#1}}%
\newcommand{\hlkwd}[1]{\textcolor[rgb]{0.737,0.353,0.396}{\textbf{#1}}}%

\usepackage{framed}
\makeatletter
\newenvironment{kframe}{%
 \def\at@end@of@kframe{}%
 \ifinner\ifhmode%
  \def\at@end@of@kframe{\end{minipage}}%
  \begin{minipage}{\columnwidth}%
 \fi\fi%
 \def\FrameCommand##1{\hskip\@totalleftmargin \hskip-\fboxsep
 \colorbox{shadecolor}{##1}\hskip-\fboxsep
     \hskip-\linewidth \hskip-\@totalleftmargin \hskip\columnwidth}%
 \MakeFramed {\advance\hsize-\width
   \@totalleftmargin\z@ \linewidth\hsize
   \@setminipage}}%
 {\par\unskip\endMakeFramed%
 \at@end@of@kframe}
\makeatother

\definecolor{shadecolor}{rgb}{.97, .97, .97}
\definecolor{messagecolor}{rgb}{0, 0, 0}
\definecolor{warningcolor}{rgb}{1, 0, 1}
\definecolor{errorcolor}{rgb}{1, 0, 0}
\newenvironment{knitrout}{}{} 

\usepackage{alltt}
\usepackage[T1]{fontenc}

\usepackage{microtype}
\usepackage{booktabs}
\usepackage{amsfonts}
\usepackage{listings}

\usepackage[%
rm={oldstyle=false,proportional=true},%
sf={oldstyle=false,proportional=true},%
tt={oldstyle=false,proportional=true,variable=true},%
qt=false%
]{cfr-lm}
\usepackage{graphicx}

\usepackage{paralist}
\usepackage[caption=false]{subfig}

\usepackage[T1]{fontenc}

\usepackage[math]{blindtext}

\usepackage{csquotes}

\usepackage{microtype}

\usepackage{url}
\makeatletter
\def\Url@twoslashes{\mathchar`\/\@ifnextchar/{\kern-.2em}{}}
\g@addto@macro\UrlSpecials{\do\/{\Url@twoslashes}}
\makeatother
\makeatletter
\g@addto@macro{\UrlBreaks}{\UrlOrds}
\makeatother

\usepackage[
bookmarks=false,
breaklinks=true,
colorlinks=true,
linkcolor=black,
citecolor=black,
urlcolor=black,
pdfpagelayout=SinglePage,
pdfstartview=Fit
]{hyperref}
\usepackage[all]{hypcap}

\usepackage[capitalise]{cleveref}
\crefname{section}{Sect.}{Sect.}
\Crefname{section}{Section}{Sections}
\crefname{figure}{Fig.}{Fig.}
\Crefname{figure}{Figure}{Figures}

\usepackage{xspace}

\DeclareFontFamily{U}{MnSymbolC}{}
\DeclareSymbolFont{MnSyC}{U}{MnSymbolC}{m}{n}
\DeclareFontShape{U}{MnSymbolC}{m}{n}{
    <-6>  MnSymbolC5
   <6-7>  MnSymbolC6
   <7-8>  MnSymbolC7
   <8-9>  MnSymbolC8
   <9-10> MnSymbolC9
  <10-12> MnSymbolC10
  <12->   MnSymbolC12%
}{}

\DeclareMathSymbol{\powerset}{\mathord}{MnSyC}{180}

\IfFileExists{upquote.sty}{\usepackage{upquote}}{}
\begin{document}

\title{RSSL: Semi-supervised Learning in R}

\author{Jesse H. Krijthe\inst{1,2}}

\institute{
Pattern Recognition Laboratory, Delft University of Technology\\
\and
Department of Molecular Epidemiology, Leiden University Medical Center\\
\email{jkrijthe@gmail.com}
}
			
\maketitle

\begin{abstract}
In this paper, we introduce a package for semi-supervised learning research in the R programming language called RSSL. We cover the purpose of the package, the methods it includes and comment on their use and implementation. We then show, using several code examples, how the package can be used to replicate well-known results from the semi-supervised learning literature.
\end{abstract}

\section{Introduction}
Semi-supervised learning is concerned with using unlabeled examples, that is, examples for which we know the values for the input features but not the corresponding outcome, to improve the performance of supervised learning methods that only use labeled examples to train a model. An important motivation for investigations into these types of algorithms is that in some applications, gathering labels is relatively expensive or time-consuming, compared to the cost of obtaining an unlabeled example. Consider, for instance, building a web-page classifier. Downloading millions of unlabeled web-pages is easy. Reading them to assign a label is time-consuming. Effectively using unlabeled examples to improve supervised classifiers can therefore greatly reduce the cost of building a decently performing prediction model, or make it feasible in cases where labeling many examples is not a viable option.

While the R programming language \cite{RCoreTeam2016} offers a rich set of implementations of a plethora of supervised learning methods, brought together by machine learning packages such as \texttt{caret} and \texttt{mlr} there are fewer implementations of methods that can deal with the semi-supervised learning setting. This both impedes the spread of the use of these types of algorithms by practitioners, and makes it harder for researchers to study these approaches or compare new methods to existing ones.  The goal of the RSSL package is to make a step towards filling this hiatus, with a focus on providing methods that exemplify common behaviours of semi-supervised learning methods.

Until recently, no package providing multiple semi-supervised learning methods was available in R\footnote{Recently, the \texttt{SSL} package was introduced whose implementations are mostly complementary to those offered in our package: \url{https://CRAN.R-project.org/package=SSL}}. In other languages, semi-supervised learning libraries that bring together several different methods are not available either, although there are general purpose machine learning libraries, such as scikit-learn in Python \cite{scikit-learn} that offer implementations of some semi-supervised algorithms. A broader set of implementations is available for Matlab, since the original implementations provided by the authors of many of the approaches covered by our package are provided for Matlab. The goal of our package is to bring some of these implementations together in the R environment by providing common interfaces to these methods, either implementing these methods in R, translating code to R or providing interfaces to C++ libraries.

The goal of this work is to give an overview of the package and make some comments how it is implemented and how it can be used. We will then provide several examples on how the package can be used to replicate various well-known results from the semi-supervised learning literature.

\section{Overview of the Package}

\subsection{Classifiers}
The package focuses on semi-supervised classification. We give an overview of the classifiers that are available in \Cref{table:classifiers}. We consider it important to compare the performance of semi-supervised learners to their supervised counterparts. We therefore include several supervised implementations and sets of semi-supervised methods corresponding to each supervised method. Most of the methods are new implementations in R based on the description of the methods in the original research papers. For others, we either provide a (close to) direct translation of the original code into R code or an R interface to the original C++ code. For the latter we make use of the Rcpp package \cite{Eddelbuettel2011}. In some cases (WellSVM and S4VM) it was necessary to also include a customized version of LIBSVM \cite{Chang2011} on which these implementations depend.

A common wrapper method for semi-supervised learning, self-learning, is available for all supervised learners, since it merely requires a supervised classifier and some unlabeled objects. Other types of semi-supervised methods that are available for multiple supervised classifiers are the moment (or intrinsically) constrained methods of \cite{Loog2010,Loog2014a}, the implicitly constrained methods of \cite{Krijthe2014,Krijthe2016,Krijthe2016d} and the Laplacian regularization of \cite{Belkin2006}.

All the classifier functions require as input either matrices with feature values (one for the labeled data and one for the unlabeled data) and a \texttt{factor} object containing the labels, or a \texttt{formula} object defining the names input and target variables and a corresponding \texttt{data.frame} object containing the whole dataset. In the examples, we will mostly use the latter style, since it fits better with the use of the pipe operator that is becoming popular in R programming.

Each classifier function returns an object of a specific subclass of the \texttt{Classifier} class containing the trained classifier. There are several methods that we can call on these objects. The \texttt{predict} method predicts the labels of new data. \texttt{decisionvalues} returns the value of the decision function for new objects. If available, the \texttt{loss} method returns the classifier specific loss (the surrogate loss used to train the classifier) incurred by the classifier on a set of examples. If the method assigns responsibilities --probabilities of belonging to a particular class-- to the unlabeled examples, \texttt{responsibilities} returns the responsibility values assigned to the unlabeled examples. For linear classifiers, we often provide the \texttt{line\_coefficients} method that provides the coefficients to plot a 2-dimensional decision boundary, which may be useful for plotting the classifier in simple 2D examples.

\begin{table}
\label{table:classifiers}
\caption{Overview of classifiers available in RSSL}
\begin{center}
\begin{tabular}{ lcccl } 
 \toprule
 \textsc{Classifier} & \textsc{R} & \textsc{Interface} & \textsc{Port} & \textsc{Reference} \\ 
 \midrule
\textbf{(Kernel) Least Squares Classifier} & \checkmark &  &  & \cite{Hastie2009} \\
\, Implicitly Constrained  & \checkmark & & & \cite{Krijthe2016d}  \\
\, Implicitly Constrained Projection  & \checkmark & & & \cite{Krijthe2016}  \\
\, Laplacian Regularized & \checkmark & & & \cite{Belkin2006} \\
\, Updated Second Moment & \checkmark & & & \cite{Shaffer1991}  \\
\, Self-learning & \checkmark & & & \cite{McLachlan1975} \\
\, Optimistic / ``Expectation Maximization'' & \checkmark & & & \cite{Krijthe2016a}  \\
\midrule
\textbf{Linear Discriminant Analysis} & \checkmark & & & \cite{Webb2002} \\
\, Expectation Maximization  & \checkmark & & & \cite{Dempster1977} \\
\, Implicitly Constrained  & \checkmark & & & \cite{Krijthe2014} \\
\, Maximum Constrastive Pessimistic  & & & \checkmark & \cite{Loog2016} \\
\, Moment Constrained  & \checkmark & & & \cite{Loog2014a} \\
\, Self-learning & \checkmark & & & \cite{McLachlan1975} \\
\midrule
\textbf{Nearest Mean Classifier} & \checkmark & & & \cite{Webb2002} \\
\, Expectation Maximization & \checkmark & & & \cite{Dempster1977} \\
\, Moment Constrained & \checkmark & & & \cite{Loog2010} \\
\, Self-learning & \checkmark & & & \cite{McLachlan1975} \\
\midrule
\textbf{Support Vector Machine} & \checkmark & & &  \\
\, SVMlin & & \checkmark & & \cite{Sindhwani2006} \\
\, WellSVM & & & \checkmark & \cite{Li2013} \\
\, S4VM & & & \checkmark & \cite{Li2015} \\
\, Transductive SVM (Convex Concave Procedure) & \checkmark & & & \cite{Joachims1999,Collobert2006} \\
\, Laplacian SVM & \checkmark & & & \cite{Belkin2006} \\
\, Self-learning & \checkmark & & & \cite{McLachlan1975} \\
\midrule
\textbf{Logistic Regression} & \checkmark & & &  \\
\, Entropy Regularized Logistic Regression & \checkmark & & & \cite{Grandvalet2005} \\
\, Self-learning & \checkmark & & & \cite{McLachlan1975} \\
\midrule
Harmonic Energy Minimization & \checkmark & & & \cite{Zhu2003} \\
\bottomrule
\end{tabular}
\end{center}
\end{table}

\subsection{Utility Functions}
In addition to the implementations of the classifiers themselves, the package includes a number of functions that simplify setting up experiments and studying these classifiers. There are three main categories of functions: functions to generate simulated datasets, functions to evaluate classifiers and run experiments and functions for plotting trained classifiers.

\subsubsection{Generated Datasets}
A number of functions, of the form \texttt{generate*}, create datasets sampled from archetypical simulated problems. An overview of simulated datasets is given in \Cref{fig:generateddatasets}. You will notice that these datasets mostly show examples where the structure of the density of the feature values is either very informative or not informative at all for the estimation of the conditional distribution of the labels given the feature value. A major theme in semi-supervised learning research is how to leverage this connection between the distribution of the features and the conditional distribution of the labels, and what happens if this connection is non-existent. These simulated datasets offer some simple but interesting test cases for semi-supervised methods.
\begin{knitrout}
\definecolor{shadecolor}{rgb}{0.969, 0.969, 0.969}\color{fgcolor}\begin{figure}
\includegraphics[width=\maxwidth]{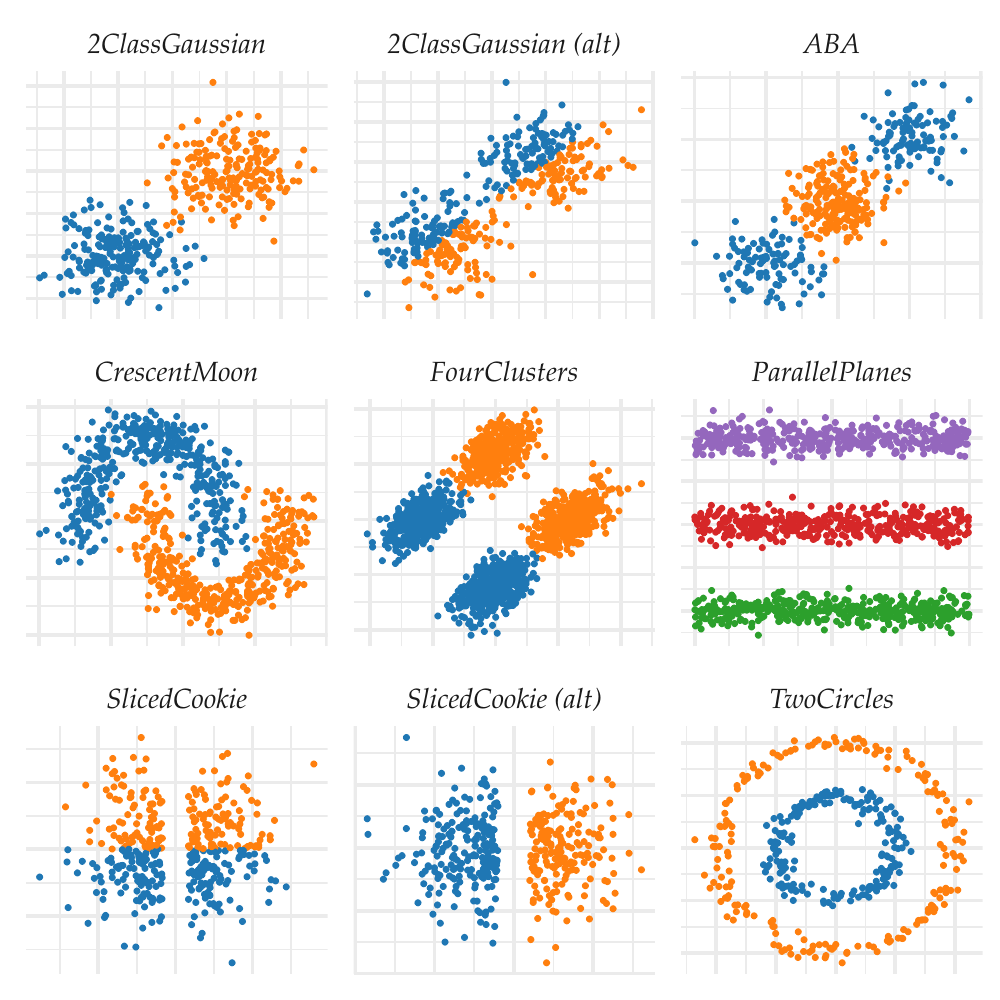} \caption[Simulated Datasets]{Simulated Datasets. Each can be generated using a function of the form \texttt{generate*}, where \texttt{*} should be replaced by the name of the dataset. (alt) indicates non-default parameters were used when calling the function.}\label{fig:generateddatasets}
\end{figure}

\end{knitrout}

\subsubsection{Classifier Evaluation}
To evaluate the performance of different methods, the package contains three types of functions that implement standard procedures for setting up such experiments. The first is by splitting a fully labeled dataset into a labeled set, an unlabeled set and a test set. For data in the form of a matrix, the \texttt{split\_dataset\_ssl} can be used. For data in the form of a data frame, the easiest way is to use \texttt{magrittr}'s pipe operator, splitting the data using the \texttt{split\_random} command, using \texttt{add\_missinglabels\_mar} to randomly remove labels, and \texttt{missing\_labels} or \texttt{true\_labels} to recover these labels when we want to evaluate the performance on the unlabeled objects.
The second type of experiment is to apply cross-validation in a semi-supervised setting using \texttt{CrossValidationSSL}. Distinct from the normal cross-validation setting, the data in the training folds get randomly assigned to the labeled or unlabeled set.
The third type of experiment enabled by the package is to generate learning curves using the \texttt{LearningCurveSSL} function. These are performance curves for increasing numbers of unlabeled examples or an increasing fraction of labeled examples. 
For both the learning curves and cross-validation, multiple datasets can be given as input and the performance measures can be user defined, or one could use one of the supplied \texttt{measure\_*} functions. Also in both cases, the experiments can optionally be run in parallel on multiple cores to speed up computation.

\subsubsection{Plotting}
Three ways to plot classifiers in simple 2D examples are provided. The most general method relies on the ggplot2 package \cite{Wickham2009} to plot the data and is provided in the form of the \texttt{stat\_classifier} that can add  classification boundaries to \texttt{ggplot2} plots. \texttt{geom\_linearclassifier} works in a similar way, but only works for a number of linear classifiers that have an associated \texttt{line\_coefficients} method. Lastly, for these classifiers \texttt{line\_coefficients} can be used directly to get the parameters that define the linear decision boundary, for use in a custom plotting function. In the examples, we will illustrate the use of \texttt{stat\_classifier} and \texttt{geom\_linearclassifier}.

\section{Installation}
The package is available from the Comprehensive R Archive Network (CRAN). As such, the easiest way to install the package is to run the following command using a recent version of R:
\begin{knitrout}
\definecolor{shadecolor}{rgb}{0.969, 0.969, 0.969}\color{fgcolor}\begin{kframe}
\begin{alltt}
\hlkwd{install.packages}\hlstd{(}\hlstr{"RSSL"}\hlstd{)}
\end{alltt}
\end{kframe}
\end{knitrout}
\noindent The latest development version of the package can be installed using:
\begin{knitrout}
\definecolor{shadecolor}{rgb}{0.969, 0.969, 0.969}\color{fgcolor}\begin{kframe}
\begin{alltt}
\hlcom{# If devtools is not installed run: install.packages("devtools")}
\hlstd{devtools}\hlopt{::}\hlkwd{install_github}\hlstd{(}\hlstr{"jkrijthe/RSSL"}\hlstd{)}
\end{alltt}
\end{kframe}
\end{knitrout}

\section{Examples}
In this section, we will provide several examples of how the RSSL package can be used to illustrate or replicate results from the semi-supervised learning literature. Due to space constraints, we provide parts of the code for the examples in the text below. The complete code for all examples can be found in the source version of this document, which can be found on the author's website\footnote{\url{www.jessekrijthe.com}}.

\subsection{A Failure of Self-Learning}
While semi-supervised learning may seem to be obviously helpful, the fact that semi-supervised methods can actually lead to worse performance than their supervised counterparts has been both widely observed and described \cite{Cozman2003}. We will generate an example where unlabeled data is helpful (using the 2ClassGaussian problem from \Cref{fig:generateddatasets}) and one where unlabeled data actually leads to an increase in the classification error (2ClassGaussian (alt) in \Cref{fig:generateddatasets}), for the least squares classifier and self-learning as the semi-supervised learner. This can be done using the following code:
\begin{knitrout}\footnotesize
\definecolor{shadecolor}{rgb}{0.969, 0.969, 0.969}\color{fgcolor}\begin{kframe}
\begin{alltt}
\hlkwd{library}\hlstd{(RSSL)}
\hlkwd{set.seed}\hlstd{(}\hlnum{1}\hlstd{)}

\hlcom{# Set the datasets and corresponding formula objects}
\hlstd{datasets} \hlkwb{<-} \hlkwd{list}\hlstd{(}\hlstr{"2 Gaussian Expected"}\hlstd{=}
                   \hlkwd{generate2ClassGaussian}\hlstd{(}\hlkwc{n}\hlstd{=}\hlnum{2000}\hlstd{,}\hlkwc{d}\hlstd{=}\hlnum{2}\hlstd{,}\hlkwc{expected}\hlstd{=}\hlnum{TRUE}\hlstd{),}
                 \hlstr{"2 Gaussian Non-Expected"}\hlstd{=}
                   \hlkwd{generate2ClassGaussian}\hlstd{(}\hlkwc{n}\hlstd{=}\hlnum{2000}\hlstd{,}\hlkwc{d}\hlstd{=}\hlnum{2}\hlstd{,}\hlkwc{expected}\hlstd{=}\hlnum{FALSE}\hlstd{))}
\hlstd{formulae} \hlkwb{<-} \hlkwd{list}\hlstd{(}\hlstr{"2 Gaussian Expected"}\hlstd{=}\hlkwd{formula}\hlstd{(Class}\hlopt{~}\hlstd{.),}
                 \hlstr{"2 Gaussian Non-Expected"}\hlstd{=}\hlkwd{formula}\hlstd{(Class}\hlopt{~}\hlstd{.))}

\hlcom{# Define the classifiers to be used}
\hlstd{classifiers} \hlkwb{<-} \hlkwd{list}\hlstd{(}\hlstr{"Supervised"} \hlstd{=}
                      \hlkwa{function}\hlstd{(}\hlkwc{X}\hlstd{,}\hlkwc{y}\hlstd{,}\hlkwc{X_u}\hlstd{,}\hlkwc{y_u}\hlstd{) \{} \hlkwd{LeastSquaresClassifier}\hlstd{(X,y)\},}
                    \hlstr{"Self-learning"} \hlstd{=}
                      \hlkwa{function}\hlstd{(}\hlkwc{X}\hlstd{,}\hlkwc{y}\hlstd{,}\hlkwc{X_u}\hlstd{,}\hlkwc{y_u}\hlstd{) \{} \hlkwd{SelfLearning}\hlstd{(X,y,X_u,}
                                                \hlkwc{method} \hlstd{= LeastSquaresClassifier)\})}

\hlcom{# Define the performance measures to be used and run experiment}
\hlstd{measures} \hlkwb{<-} \hlkwd{list}\hlstd{(}\hlstr{"Error"} \hlstd{=  measure_error,} \hlstr{"Loss"} \hlstd{= measure_losstest)}
\hlstd{results_lc} \hlkwb{<-} \hlkwd{LearningCurveSSL}\hlstd{(formulae,datasets,}
                           \hlkwc{classifiers}\hlstd{=classifiers,}
                           \hlkwc{measures}\hlstd{=measures,}\hlkwc{verbose}\hlstd{=}\hlnum{FALSE}\hlstd{,}
                           \hlkwc{repeats}\hlstd{=}\hlnum{100}\hlstd{,}\hlkwc{n_l}\hlstd{=}\hlnum{10}\hlstd{,}\hlkwc{sizes} \hlstd{=} \hlnum{2}\hlopt{^}\hlstd{(}\hlnum{1}\hlopt{:}\hlnum{10}\hlstd{))}
\end{alltt}
\end{kframe}
\end{knitrout}
\noindent When we plot these results (using the \texttt{plot} method and optionally changing the display settings of the plot), we get the figure shown in \Cref{fig:plot-lc}. What this shows is that, clearly, semi-supervised methods can be outperformed by their supervised counterpart for some datasets, for some choice of semi-supervised learner. Given that one may have little labeled training data to accurately detect that this is happening, in some settings we may want to consider methods that inherently attempt to avoid this deterioration in performance. We will return to this in a later example.

\begin{knitrout}
\definecolor{shadecolor}{rgb}{0.969, 0.969, 0.969}\color{fgcolor}\begin{figure}
\includegraphics[width=\maxwidth]{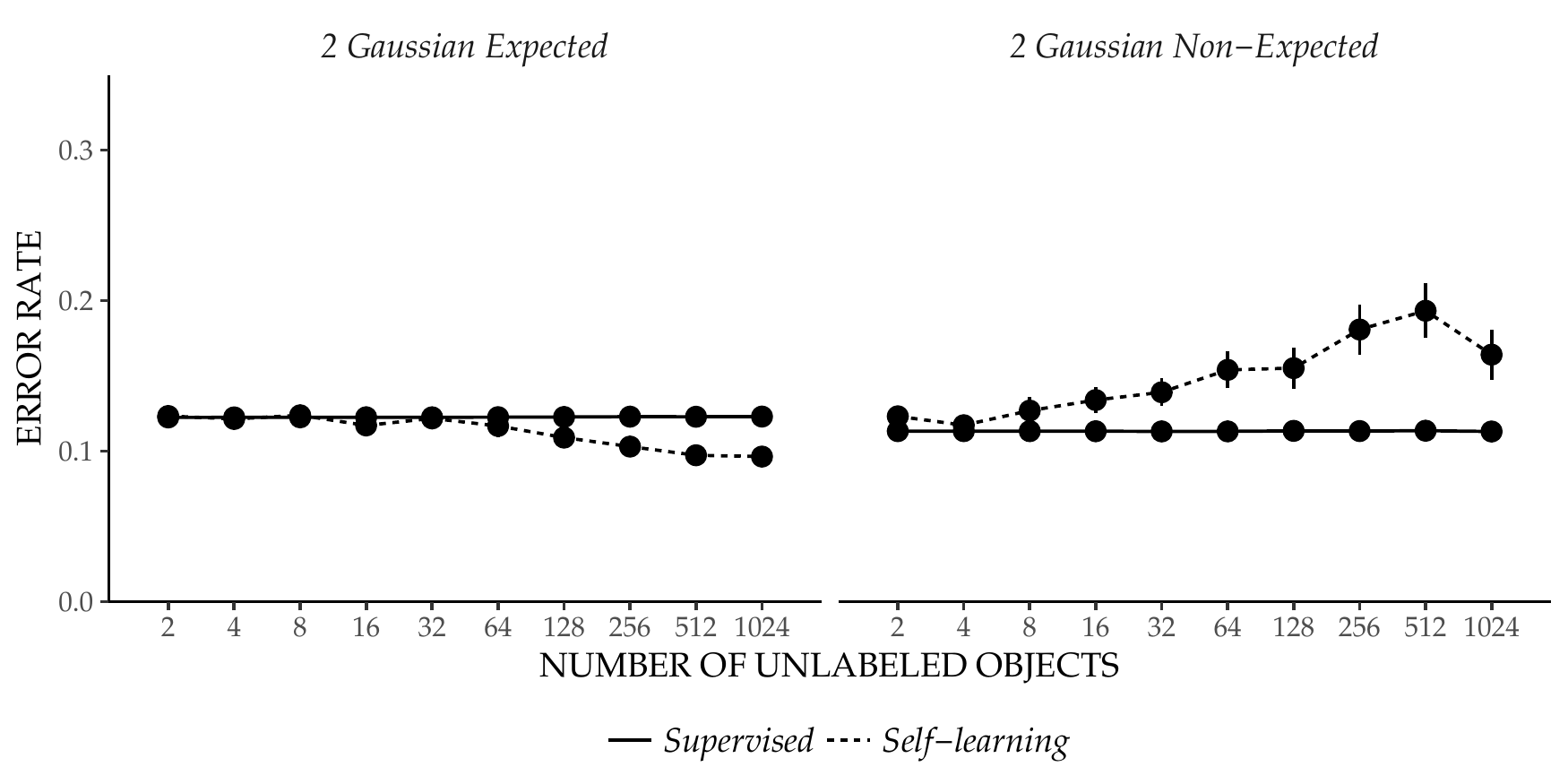} \caption{Example where self-learning leads to better performance as we add more unlabeled data (left) and increasingly worse performance as unlabeled data is added (right). The classifier used is the least squares classifier. The datasets are similar to the ones shown in \Cref{fig:generateddatasets}.}\label{fig:plot-lc}
\end{figure}

\end{knitrout}

\subsection{Graph Based Semi-supervised Learning}
Many methods in semi-supervised learning attempt to use the assumption that labels change smoothly over dense regions in the feature space. An early attempt to encode this assumption is offered by \cite{Zhu2003} who propose to minimize an energy function for the labels of the unlabeled objects that penalizes large deviations between labels assigned to objects that are close, for some measure of closeness. This so-called harmonic energy formulation can also be interpreted as a propagation of the labels from the labeled objects to the unlabeled objects, through a graph that encodes a measure of closeness. We recreate \cite{Zhu2003}'s Figure~2, which can be found in \Cref{fig:labelpropagation}. Due to space constraints, we will defer the code to the online version of this document, since it is similar to the code for the next example.
\begin{knitrout}
\definecolor{shadecolor}{rgb}{0.969, 0.969, 0.969}\color{fgcolor}\begin{figure}
\subfloat[Parallel planes dataset\label{fig:labelpropagation1}]{\includegraphics[width=.48\linewidth]{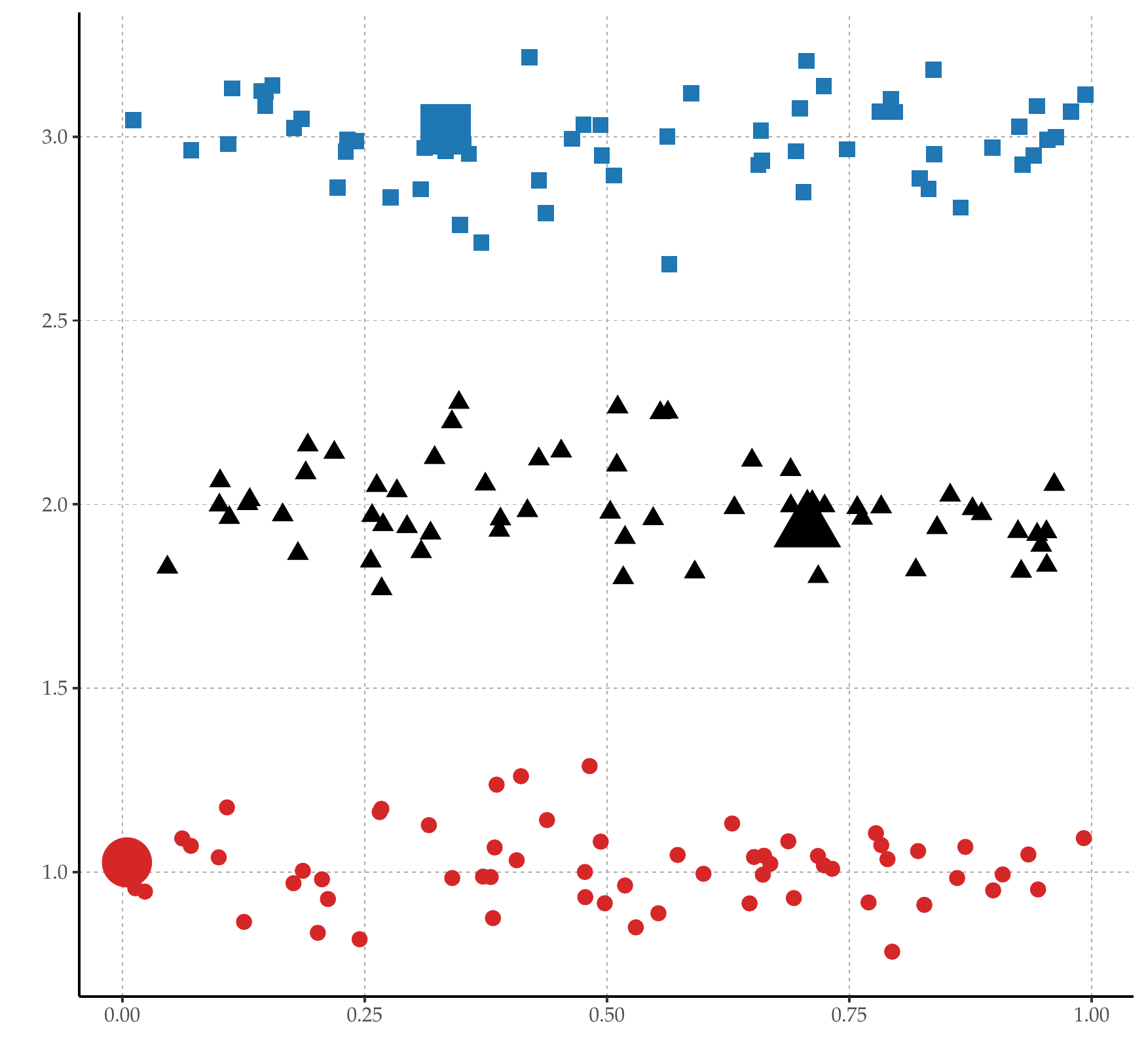} }
\subfloat[Spirals dataset\label{fig:labelpropagation2}]{\includegraphics[width=.48\linewidth]{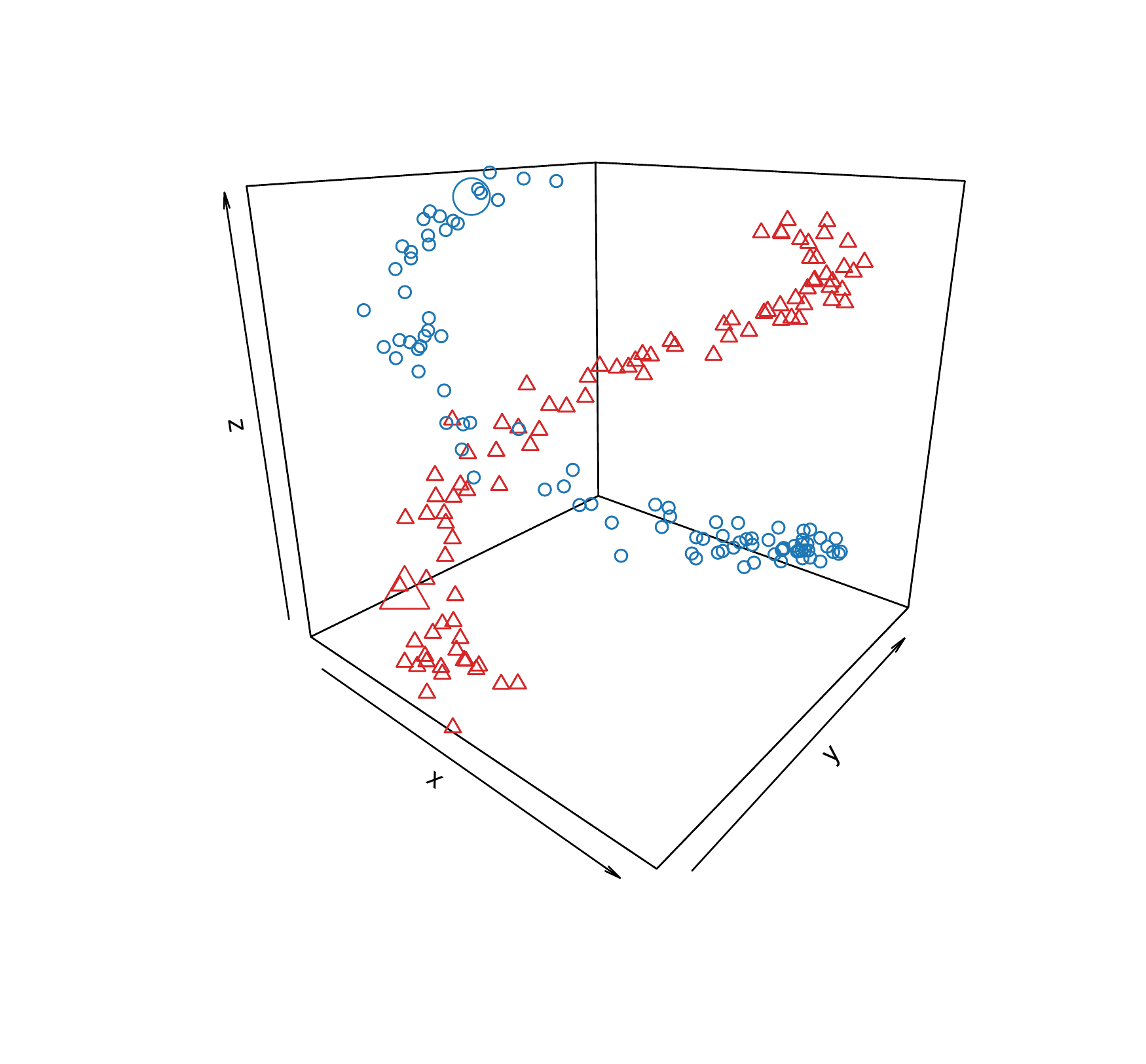} }\caption[Replication of Figure 2 from \cite{Zhu2003} demonstrating harmonic energy minimization]{Replication of Figure 2 from \cite{Zhu2003} demonstrating harmonic energy minimization. The larger points indicate the labeled objects. The color indicates the predicted class.}\label{fig:labelpropagation}
\end{figure}

\end{knitrout}

\subsection{Manifold Regularization}
Belkin et al. \cite{Belkin2006} build on the ideas of \cite{Zhu2003} by formulating the smoothness of the labeling function over the data manifold as a regularization term. In RSSL this Laplacian regularization term is included in both an SVM formulation and a regularized least squares formulation. For the Laplacian SVM formulation, Figure~2 from \cite{Belkin2006} provides an example of its performance on a simulated dataset. We can replicate this result using the following code. The results are shown in \Cref{fig:manifoldregularization}.
\begin{knitrout}\footnotesize
\definecolor{shadecolor}{rgb}{0.969, 0.969, 0.969}\color{fgcolor}\begin{kframe}
\begin{alltt}
\hlkwd{library}\hlstd{(RSSL)}
\hlkwd{library}\hlstd{(dplyr)}
\hlkwd{library}\hlstd{(ggplot2)}
\hlstd{plot_style} \hlkwb{<-} \hlkwd{theme_classic}\hlstd{()} \hlcom{# Set the style of the plot}

\hlkwd{set.seed}\hlstd{(}\hlnum{2}\hlstd{)}
\hlstd{df_unlabeled} \hlkwb{<-} \hlkwd{generateCrescentMoon}\hlstd{(}\hlkwc{n}\hlstd{=}\hlnum{100}\hlstd{,}\hlkwc{sigma} \hlstd{=} \hlnum{0.3}\hlstd{)} \hlopt{%>%}
  \hlkwd{add_missinglabels_mar}\hlstd{(Class}\hlopt{~}\hlstd{.,}\hlkwc{prob}\hlstd{=}\hlnum{1}\hlstd{)}
\hlstd{df_labeled} \hlkwb{<-} \hlkwd{generateCrescentMoon}\hlstd{(}\hlkwc{n}\hlstd{=}\hlnum{1}\hlstd{,}\hlkwc{sigma} \hlstd{=} \hlnum{0.3}\hlstd{)}
\hlstd{df} \hlkwb{<-} \hlkwd{rbind}\hlstd{(df_unlabeled,df_labeled)}

\hlstd{c_svm} \hlkwb{<-} \hlkwd{SVM}\hlstd{(Class}\hlopt{~}\hlstd{.,df_labeled,}\hlkwc{scale}\hlstd{=}\hlnum{FALSE}\hlstd{,}
             \hlkwc{kernel} \hlstd{= kernlab}\hlopt{::}\hlkwd{rbfdot}\hlstd{(}\hlnum{0.05}\hlstd{),}
             \hlkwc{C}\hlstd{=}\hlnum{2500}\hlstd{)}

\hlstd{c_lapsvm1} \hlkwb{<-} \hlkwd{LaplacianSVM}\hlstd{(Class}\hlopt{~}\hlstd{.,df,}\hlkwc{scale}\hlstd{=}\hlnum{FALSE}\hlstd{,}
                         \hlkwc{kernel}\hlstd{=kernlab}\hlopt{::}\hlkwd{rbfdot}\hlstd{(}\hlnum{0.05}\hlstd{),}
                         \hlkwc{lambda} \hlstd{=} \hlnum{0.0001}\hlstd{,}\hlkwc{gamma}\hlstd{=}\hlnum{10}\hlstd{)}

\hlstd{c_lapsvm2} \hlkwb{<-} \hlkwd{LaplacianSVM}\hlstd{(Class}\hlopt{~}\hlstd{.,df,}\hlkwc{scale}\hlstd{=}\hlnum{FALSE}\hlstd{,}
                         \hlkwc{kernel}\hlstd{=kernlab}\hlopt{::}\hlkwd{rbfdot}\hlstd{(}\hlnum{0.05}\hlstd{),}
                         \hlkwc{lambda} \hlstd{=} \hlnum{0.0001}\hlstd{,}\hlkwc{gamma}\hlstd{=}\hlnum{10000}\hlstd{)}

\hlcom{# Plot the results }
\hlcom{# Change the arguments of stat_classifier to plot the Laplacian SVM}
\hlkwd{ggplot}\hlstd{(df_unlabeled,} \hlkwd{aes}\hlstd{(}\hlkwc{x}\hlstd{=X1,}\hlkwc{y}\hlstd{=X2))} \hlopt{+}
  \hlkwd{geom_point}\hlstd{()} \hlopt{+}
  \hlkwd{geom_point}\hlstd{(}\hlkwd{aes}\hlstd{(}\hlkwc{color}\hlstd{=Class,}\hlkwc{shape}\hlstd{=Class),}\hlkwc{data}\hlstd{=df_labeled,}\hlkwc{size}\hlstd{=}\hlnum{5}\hlstd{)} \hlopt{+}
  \hlkwd{stat_classifier}\hlstd{(}\hlkwc{classifiers}\hlstd{=}\hlkwd{list}\hlstd{(}\hlstr{"SVM"}\hlstd{=c_svm),}\hlkwc{color}\hlstd{=}\hlstr{"black"}\hlstd{)} \hlopt{+}
  \hlkwd{ggtitle}\hlstd{(}\hlstr{"SVM"}\hlstd{)}\hlopt{+}
  \hlstd{plot_style}
\end{alltt}
\end{kframe}\begin{figure}
\subfloat[$\lambda=0.0001$, $\gamma=0$\label{fig:manifoldregularization1}]{\includegraphics[width=.33\linewidth]{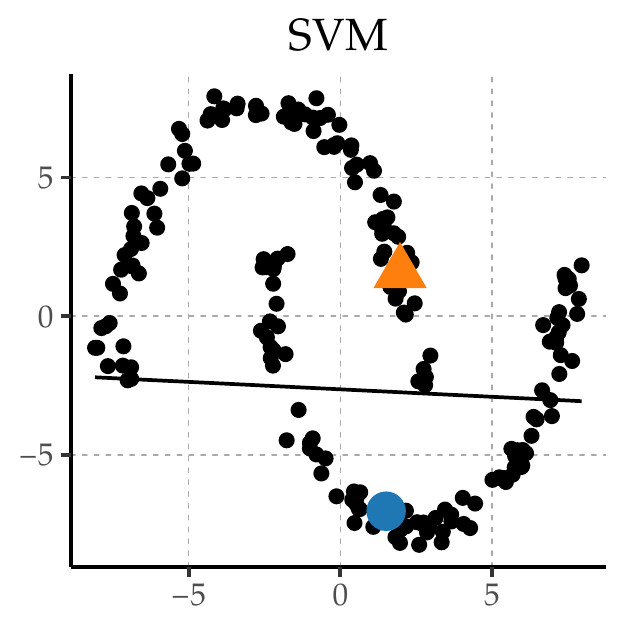} }
\subfloat[$\lambda=0.0001$, $\gamma=10$\label{fig:manifoldregularization2}]{\includegraphics[width=.33\linewidth]{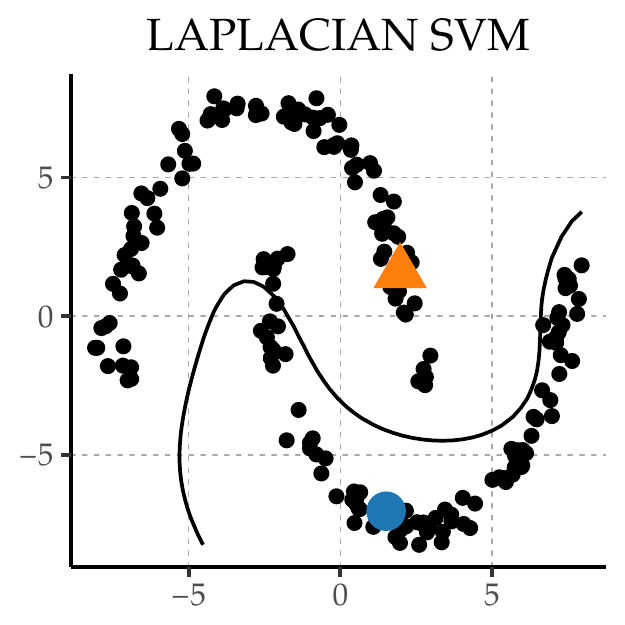} }
\subfloat[$\lambda=0.0001$, $\gamma=10000$\label{fig:manifoldregularization3}]{\includegraphics[width=.33\linewidth]{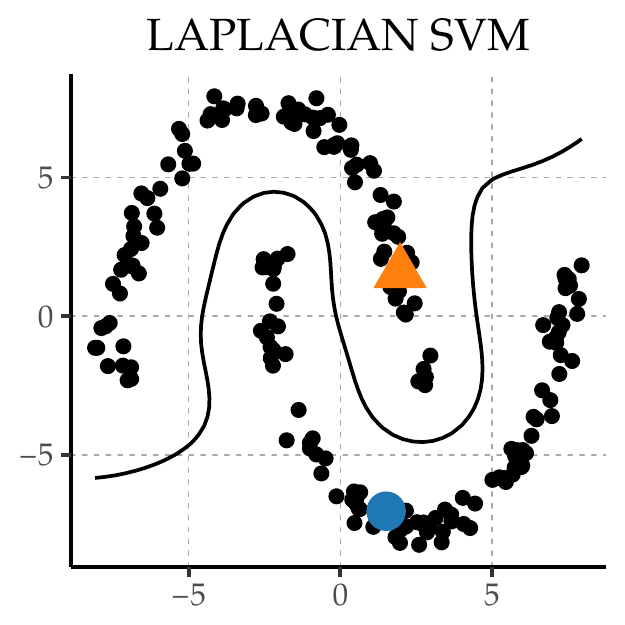} }\caption[Replication of Figure~2 from \cite{Belkin2006}]{Replication of Figure~2 from \cite{Belkin2006}. Laplacian SVM for various values of the influence of the unlabeled data.}\label{fig:manifoldregularization}
\end{figure}

\end{knitrout}

\subsection{Low Density Separation}
A number of semi-supervised approaches attempt to leverage the assumption that the classification boundary may reside in a region of low-density. The Semi-supervised SVM or Transductive SVM \cite{Joachims1999} is one such approach. In \cite[Chapter 6]{Zhu2009}, an example is given for the potential problems this low-density assumption may cause when it is not valid by considering two artificial datasets. Here we replicate these results for a different classifier that makes use of the low-density assumption: entropy regularized logistic regression \cite{Grandvalet2005}. The results are shown in \Cref{fig:lowdensityproblem}. The code to generate these results can be found in the source version of this document.
\begin{knitrout}\footnotesize
\definecolor{shadecolor}{rgb}{0.969, 0.969, 0.969}\color{fgcolor}\begin{figure}
\subfloat[Low-density assumption\label{fig:lowdensityproblem1}]{\includegraphics[width=.49\linewidth]{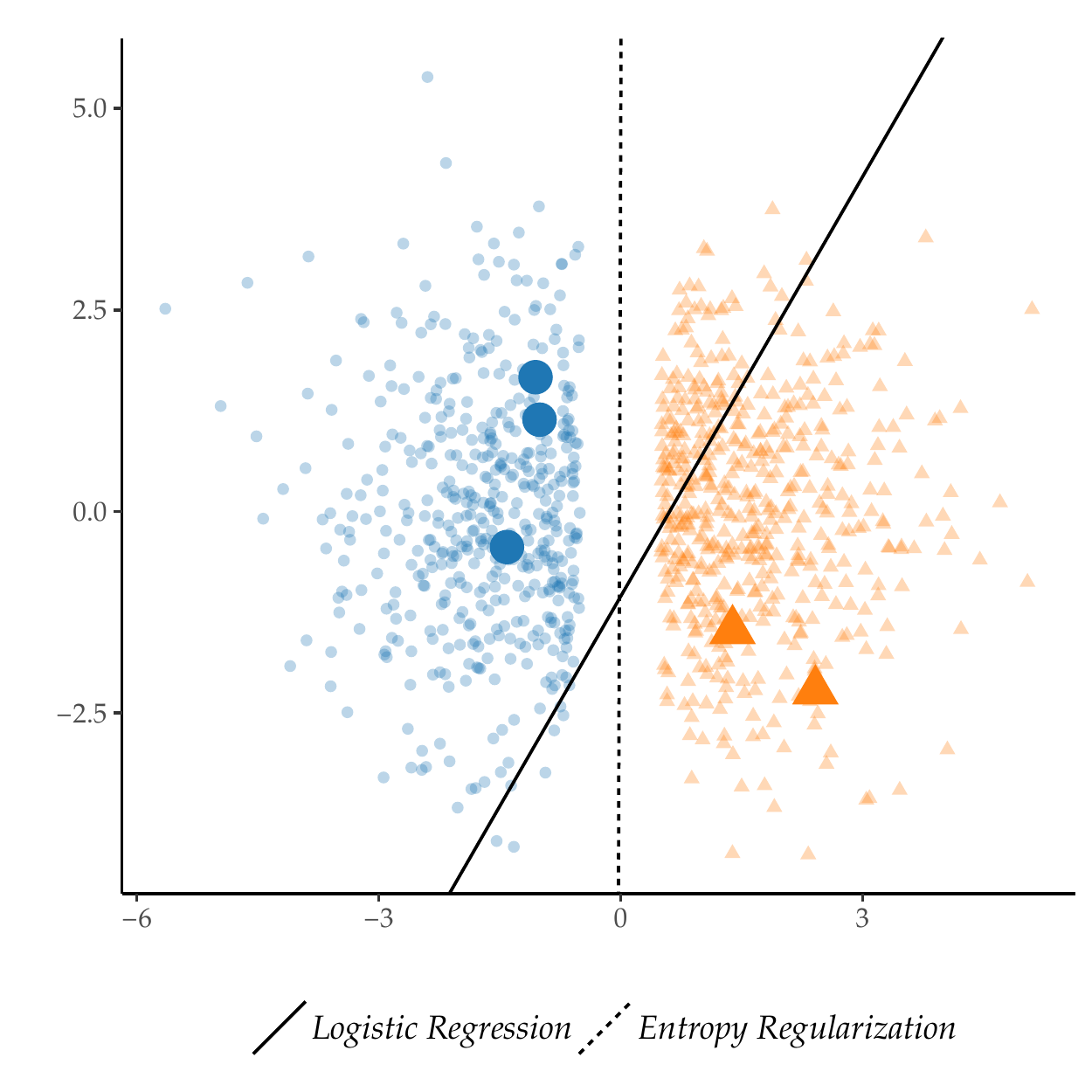} }
\subfloat[Non low-density assumption\label{fig:lowdensityproblem2}]{\includegraphics[width=.49\linewidth]{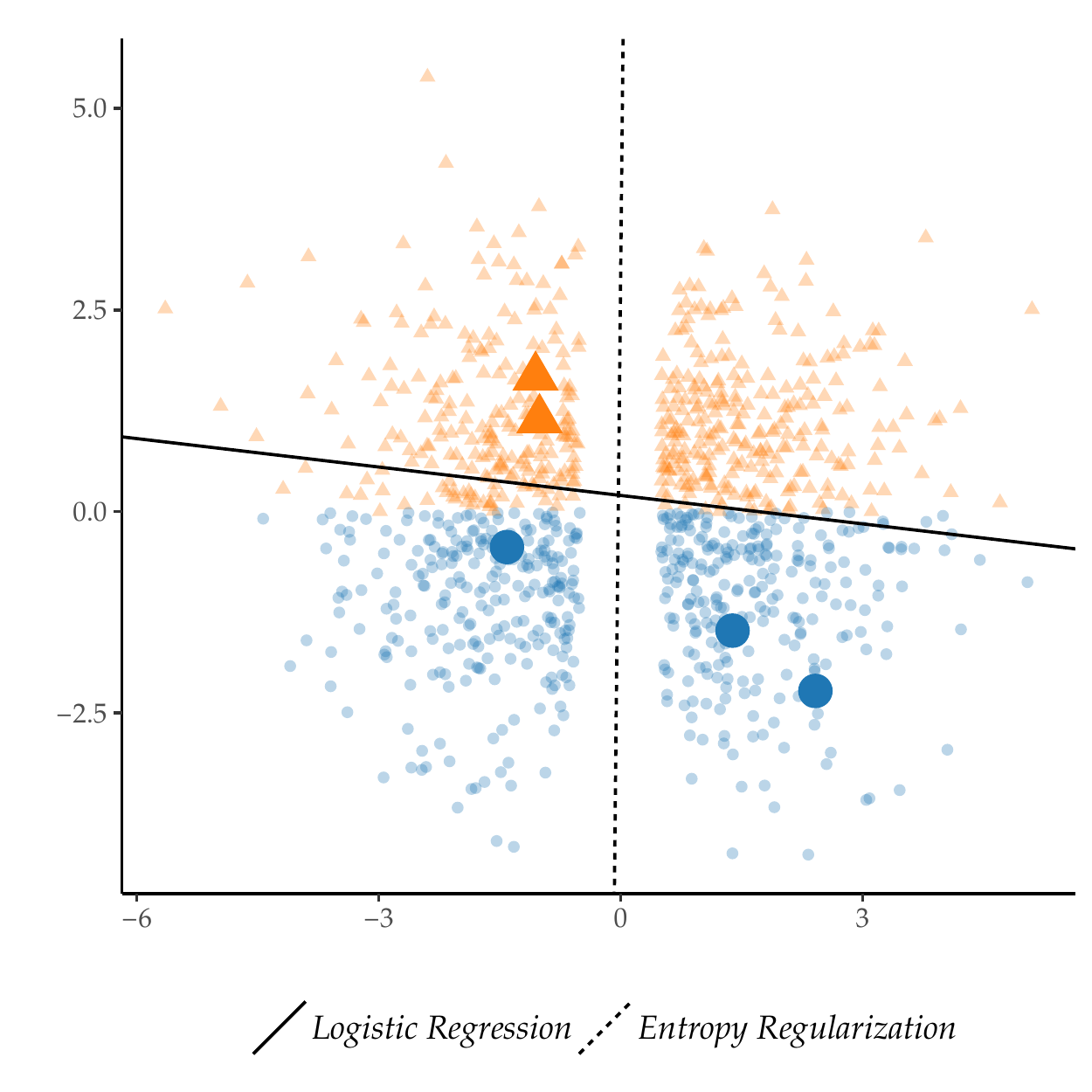} }\caption[Demonstration of potential problems when the low density assumption does not hold, similar to Figure 6.5 in \cite{Zhu2009}]{Demonstration of potential problems when the low density assumption does not hold, similar to Figure 6.5 in \cite{Zhu2009}.}\label{fig:lowdensityproblem}
\end{figure}

\end{knitrout}

\subsection{Improvement Guarantees}
We now return to the example of deterioration in performance from \Cref{fig:plot-lc}. The goal of our work in \cite{Loog2016,Krijthe2016a,Krijthe2016} is to construct methods that are guaranteed to outperform the supervised alternative. The guarantee that is given in these works is that the semi-supervised learner outperforms the supervised learner on the full, labeled and unlabeled, training set in terms of the surrogate loss (cf. \cite{Loog2014b}). The following code trains semi-supervised classifiers in these cases and returns the mean loss on the whole training set, the output is shown below the code example. It shows that indeed, these methods do not deteriorate performance in terms of the surrogate loss, while the self-learning method does show this deterioration in performance.
\begin{knitrout}\footnotesize
\definecolor{shadecolor}{rgb}{0.969, 0.969, 0.969}\color{fgcolor}\begin{kframe}
\begin{alltt}
\hlkwd{library}\hlstd{(RSSL)}
\hlkwd{set.seed}\hlstd{(}\hlnum{1}\hlstd{)}

\hlcom{# Generate Example}
\hlstd{df} \hlkwb{<-} \hlkwd{generate2ClassGaussian}\hlstd{(}\hlkwc{n}\hlstd{=}\hlnum{1000}\hlstd{,} \hlkwc{d}\hlstd{=}\hlnum{2}\hlstd{,} \hlkwc{expected}\hlstd{=}\hlnum{FALSE}\hlstd{)}
\hlstd{df_semi} \hlkwb{<-} \hlkwd{add_missinglabels_mar}\hlstd{(df, Class}\hlopt{~}\hlstd{.,} \hlkwc{prob}\hlstd{=}\hlnum{0.995}\hlstd{)}

\hlcom{# Train and evaluate classifiers}
\hlkwd{mean}\hlstd{(}\hlkwd{loss}\hlstd{(}\hlkwd{LeastSquaresClassifier}\hlstd{(Class}\hlopt{~}\hlstd{.,df_semi),df))}
\hlkwd{mean}\hlstd{(}\hlkwd{loss}\hlstd{(}\hlkwd{SelfLearning}\hlstd{(Class}\hlopt{~}\hlstd{.,df_semi,}\hlkwc{method}\hlstd{=LeastSquaresClassifier),df))}
\hlkwd{mean}\hlstd{(}\hlkwd{loss}\hlstd{(}\hlkwd{ICLeastSquaresClassifier}\hlstd{(Class}\hlopt{~}\hlstd{.,df_semi),df))}
\hlkwd{mean}\hlstd{(}\hlkwd{loss}\hlstd{(}\hlkwd{ICLeastSquaresClassifier}\hlstd{(Class}\hlopt{~}\hlstd{.,df_semi,}
                          \hlkwc{projection}\hlstd{=}\hlstr{"semisupervised"}\hlstd{),df))}
\end{alltt}
\begin{verbatim}
## [1] 0.1763921
## [1] 0.4813863
## [1] 0.1185772
## [1] 0.1236701
\end{verbatim}
\end{kframe}
\end{knitrout}

\section{Conclusion}
We presented RSSL, a package containing implementations and interfaces to implementations of semi-supervised classifiers, and utility methods to carry out experiments using these methods. We demonstrated how the package can be used to replicate several results from the semi-supervised learning literature. More usage examples can be found in the package documentation. We hope the package inspires practitioners to consider semi-supervised learning in their work and we invite others to contribute to and use the package for research. Moreover, we hope the package contributes towards making semi-supervised learning research, and the research of those who use these methods in an applied setting, fully reproducible.

\subsubsection*{Acknowledgements.}
This work was funded by project P23 of the Dutch public/private research network COMMIT.

\bibliographystyle{splncs03}
\bibliography{library}
\end{document}